\documentclass{article}
\usepackage{graphicx}
\usepackage{nips00e}
\usepackage{amsgen,amssymb,amsopn,amsmath}
\usepackage[square]{natbib}

 \title{Variable and Fixed Interval Exponential Smoothing}
\author{ Javier R. Movellan\\University of California San Diego}

\empty

\begin{document}
\maketitle

\newpage	
\section{Introduction}
Exponential smoothers provide an efficient way to compute moving
averages of signals \cite{Tong90,Box2004,Luktepohl2010}. This can be
particularly useful for real time applications. Here we define,
describe and analyze exponential smoothing algorithms, with an eye
towards practical applications.

Let $X_{t_1},X_{t_2},\cdots $ be a sequence of random variables observed at times
$t_1\leq t_2\leq t_3\leq\cdots$ be the times at which the variables are
observed. We define the exponentially smoothed average $\hat X_t$
evaluated at time $t$ as follows

\begin{align}
\hat X_t &= \frac{\tilde X_t}{\tilde w_t}\label{eqn:hatx}\\
\tilde X_t &= \sum_{t_k \leq t} e^{-(t-t_k)/\tau} X_k \\
\tilde w_t &= \sum_{t_k \leq t} e^{-(t-t_k)/\tau} 
\end{align}
where $\tau \geq 0$ is known as the time scale parameter. 

\section{Recursive Update Equations}
A key advantage of exponential smoothing is that it does not require storage of  past observations in memory. The current smoothed average, the current new observation, the time of the last observation and the time of the current observation determine the new smoothed value.  
\paragraph{Version 1:}
Note
\begin{align}
\tilde X_{t_k} = X_{t_k} + e^{-\delta(k,k-1)/\tau} X_{t_{k-1}} +  e^{-\delta(k,k-2)/\tau} X_{t_{k-2}} + \cdots + e^{-\delta(k,1)/\tau} X_{t_{1}}
\end{align}
where $\delta(k,j) = t_k -t_j$. Moreover
\begin{align}
\tilde X_{t_{k-1}} = X_{t_{k-1}} + e^{-\delta(k-1,k-2)/\tau} X_{t_{k-2}} + \cdots+ e^{-\delta(k-1,1)/\tau} X_{t_{1}}
\end{align}
Thus
\begin{align}
&\tilde X_{t_k} = X_{t_k} + \alpha_{t_k} \tilde X_{t_{k-1}}\\
&\alpha_{t_k} = e^{-\delta(k,k-1)/\tau}
\end{align}
with initial condition $\hat X_{t_1} = X_1$. Similarly
\begin{align}
\tilde w_{t_k} = \tilde w_{t_k} + \alpha_{t_k} 
\end{align}
with initial condition $\tilde w_{t_1} = 1$.  If $t$ is not one of the
sampling times, i.e., $t \in (t_{k}, t_{k+1})$ then
\begin{align}
\bar X_t & = e^{-(t-t_k)/\tau} \tilde X_{t_k}\\
\tilde w_t &= e^{-(t-t_k)/\tau} \tilde w_{t_k}\\
\hat X_t &= \frac{\tilde X_t}{\tilde w_t} = \hat X_{t_k}
\end{align}

\paragraph{Version 2;}
We shall see later that the previous update equation may run into
numerical problems as $\alpha \to 0$. Here we see a version of the
update equations that avoids this problem, though it runs into other
issues as $\alpha \to 1$. The trick is to multiply the numerator and
denominator of \eqref{eqn:hatx} by $1-\alpha$, which is acceptable as
long as $\alpha <1$. Regarding the numerator of \eqref{eqn:hatx}, let
\begin{align}
\bar X_{t_k} &= (1- \alpha_{t_k} ) \tilde X_{t_k} =
(1- \alpha_{t_k} )  X_{t_k} \nonumber\\
&+ (1- \alpha_{t_k} ) \alpha_{t_k} \tilde X_{t_{k-1}}\nonumber \\
&=(1- \alpha_{t_k})  X_{t_k}  +  \alpha_{t_k}  \bar X_{t_{k-1}} \\
&= X_{t_k} + (1-\alpha_{t_k}) (X_{t_k} - \bar X_{t_{k-1}})
\end{align}
Regarding the denominator of \eqref{eqn:hatx} let
\begin{align}
\bar w_{t_k} =& (1- \alpha_{t_k}) \tilde w_{t_k} =
(1- \alpha_{t_k}  ) + (1- \alpha_{t_k}  ) \alpha_{t_k} \tilde w_{t_{k-1}}\nonumber \\
=&(1- \alpha_{t_k} )   +  \alpha_{t_k}  \bar w_{t_{k-1}} \nonumber\\
=& 1 + (1-\alpha_{t_k}) (1 - \bar w_{t_{k-1}})
\end{align}
Then
\begin{align}
\hat X_{t_k} =\frac{\tilde X_{t_k}}{\tilde w_{t_k}}  =\frac{(1- \alpha_{t_k} ) \tilde  X_t}{(1- \alpha_{t_k} )  \tilde w_t} = \frac{\bar X_{t_k}}{\bar w_{t_k}}
\end{align}

\section{Constant Sampling Rate}
It is useful to analyze how the update equations behave when the
sampling rate is constant. In this case the time interval between
observations is a constant $\delta$ and thus $\alpha =
e^{-\delta/\tau}$ is also constant.
\paragraph{Update Equations Version 1:} Consider update equations of the form
\begin{align}
u_{k+1} = c + \alpha u_k
\end{align}
where $c$ is a constant. At equilibrium 
\begin{align}
&u_k = c + \alpha u_k \\
&u_k = \frac{c}{1-\alpha}
\end{align}
Thus at equilibrium 
\begin{align}
&\tilde w_{t_k} =  1 + \alpha \tilde w_{t_{k-1}}\\
&\tilde w_{t_k}  = \frac{1}{1-\alpha}
\end{align}
Note that $w_{t_k}$ can grow very large for $\alpha \to 1$, potentially causing numerical problems.
If the $X_{t_k}$ variables are independent identically distributed random
variables with mean $\mu$ and variance $\sigma^2$ then 
the equilibrium equations for the expected value are as follow:
\begin{align}
&E[\tilde  X_{t_k}] =  \mu + \alpha E[X_{t_k}] \\
&E[\tilde X_{t_k}] = \frac{\mu}{1-\alpha}\\
&E[\hat X_{t_k}] = \frac{E[\bar X_{t_k}]}{\bar w_{t_k} } = \mu
\end{align}
and the  equilibrium equations for the variance are as follow
\begin{align}
&Var[\tilde  X_{t_k}] =  \sigma^2  + \alpha^2  Var[X_{t_k}] \\
&Var[\tilde  X_{t_k}] =  \frac{\sigma^2}{1 - \alpha^2}   \\
&Var[\hat X_{t_k}] = \frac{(1-\alpha)^2}{1-\alpha^2} \sigma^2 = \frac{1-\alpha}{1+\alpha} \sigma^2
\end{align}
\paragraph{Update equations Version 2:}
In this case 
\begin{align}
&\bar X_{t_k} =(1-\alpha)  X_{t_k} + \alpha \bar X_{t_{k-1}}= X_{t_k} + (1-\alpha) (X_{t_k} - \bar X_{t_{k-1}}) \\
&\bar w_{t_k} = (1-\alpha) + \alpha \bar w_{t_{k-1}}
= 1 + (1-\alpha)(1 - \bar w_{t_{k-1}})
\end{align}
The equilibrium equation  for $\bar w$ is as follows
\begin{align}
\bar w_{t_k} = (1-\alpha) + \alpha \bar w_{t_k}\\
\end{align}
Thus at equilibrium
\begin{align}
\bar w_{t_k} =1
\end{align}
This can be also seen using  the sum of geometric series
\begin{equation}
\alpha^0 + \alpha^1 + \alpha^2 + \cdots + \alpha^{n-1} = \frac{1- \alpha^n}{1 - \alpha}
\end{equation}
Thus
\begin{align}
&\bar w_{t_k} = 1 - \alpha^k\\
&\lim_{k\to \infty} \bar w_{t_k}= 1
\end{align}
Thus, asymptotically, as $t_k \to \infty$ 
\begin{align}
&\hat X_{t_k} = \alpha X_{t_k} + (1-\alpha) \bar X_{t_{k-1}} =
X_{t_k} + (1-\alpha)(X_{t_k} - \hat X_{t_{k-1}})
\end{align}
If the observations are independent identically distributed with mean $\mu$ and variance $\sigma^2$ then the  asymptotic equations for the mean and variance of the smoother are as follow
\begin{align}
&E[\hat X_t] = \alpha \mu + (1 -\alpha) E[\hat X_t]\\
&Var[\hat X_t] = \alpha^2 \sigma^2 +(1-\alpha^2) Var[\hat X_t]
\end{align}
Thus, asymptotically, as $t_k \to \infty$ 
\begin{align}
&\text{E}[\hat X_{t_{k}}] = \frac{\alpha \mu}{ \alpha} = \mu\\
&\text{Var}[ \hat X_{t_{k}}]=\frac{(1-\alpha)^2}{1-\alpha^2}\sigma^2 = \frac{1-\alpha}{1+\alpha}\sigma^2
\end{align}

\paragraph{Effective Number of Averaged Observations:}
If  $\hat X$ were the average of $n$ independent observations  the variance of $\hat X$  would be $\sigma^2/n$. Thus
asymptotically, the effective number of observations averaged by the
exponential smoother can be defined as follows

\begin{align}
n = \frac{1+\alpha}{1-\alpha}\\
\alpha = \frac{n-1}{n+1}
\end{align}
Thus, for $\alpha=0$ we average one observation. The number of
averaged observations increases unboundedly as $\alpha$ increases.
\paragraph{Effective Time Window:}
In a time window of size $T$ we get $T/\delta$ observations. Thus
\begin{align}
\alpha = e^{-\delta/\tau} = \frac{T/\delta -1}{T/\delta +1} = \frac{T-\delta}{T+\delta}
\end{align}
Taking logs
\begin{align}
-\frac{\delta}{\tau} = \log(T-\delta) - \log(T+\delta)\\
\frac{1}{\tau} = \log (T+\delta) - \log(T-\delta)\\
\frac{1}{\tau} = \frac{\log (T+\delta) - \log(T-\delta)}{\delta}
\end{align}
Thus, in the limit as $\delta \to 0$
\begin{align}
&\frac{1}{\tau}= 2 \frac{d \log(T)}{dT} = \frac{2}{T}\\
&\tau  =\frac{T}{2}
\end{align}
i.e., the effective window size $T$ is twice the time scale parameter $\tau$. 
Turns out for $\delta <0.1$, the approximation
\begin{align}
\frac{\log (T+\delta) -\log(T-\delta)} {\delta} \approx \frac{2}{T}
\end{align}
is quite good. Under this approximation we get
\begin{align}
&\delta/\tau = \frac{2\delta}{T} = \frac{2}{n}\\
&\alpha = e^{-\delta/\tau}=e^{-2/n}\\
&n = 2 \log(\alpha^{-1} )
\end{align}

\section{Alternative Approaches}
Kalman filters provide an alternative way to deal with variable
interval observations. In the constant interval case Kalman filters
and exponential smoothers are asymptotically equivalent. In the
variable interval case they both weight less the last smooth value the
longer the time between the last observation and the current
observation. However they differ in the way they do so. An advantage
of exponential smoothers over Kalman filters is their simplicity. An
advantage of Kalman filters is that they provide estimates of the
uncertainty of the smoothed values. This can be useful in some
situations.

\bibliographystyle{plainnat} 
\bibliography{expSmooth}

\end{document}